\documentclass{article} % For LaTeX2e
\usepackage{fullpage}
\usepackage{cite}

\usepackage{amsmath,amsfonts,bm}

\def\eqref#1{equation~\ref{#1}}
\def\1{\bm{1}}

\DeclareMathAlphabet{\mathsfit}{\encodingdefault}{\sfdefault}{m}{sl}
\SetMathAlphabet{\mathsfit}{bold}{\encodingdefault}{\sfdefault}{bx}{n}

\usepackage{hyperref}
\usepackage{url}

\usepackage{graphicx}

\title{Symbolically Solving Partial Differential Equations using Deep Learning}

\author{Maysum Panju, Kourosh Parand, Ali Ghodsi}
\begin{document}

\maketitle

\begin{abstract}
We describe a neural-based method for generating exact or approximate solutions to differential equations in the form of mathematical expressions. Unlike other neural methods, our system returns symbolic expressions that can be interpreted directly. Our method uses a neural architecture for learning mathematical expressions to optimize a customizable objective, and is scalable, compact, and easily adaptable for a variety of tasks and configurations. The system has been shown to effectively find exact or approximate symbolic solutions to various differential equations with applications in natural sciences. In this work, we highlight how our method applies to partial differential equations over multiple variables and more complex boundary and initial value conditions. 
\end{abstract}

\section{Introduction}

Much of the physics governing the natural world can be written in the language of differential equations. Despite their simple appearances, these equations can be as challenging to solve as they are common. With recent advancements in machine learning, and deep learning in particular, a new tool became available for finding solutions to problems that previously had seemed impenetrable.

The typical approach that neural networks take for solving differential equations is to model the solution using a neural network function itself. This network is trained to fit the differential equation and produces highly accurate approximations to how the solution is supposed to behave. Although this is practically useful, it does not have the advantage of  clarity and conciseness that symbolic mathematical expressions provide. Unfortunately, the application of deep learning to produce symbolic solutions is relatively underdeveloped.

In this work, we present a method for generating symbolic solutions to differential equations that takes advantage of the flexibility and strength of deep learning. Our method leverages the power of gradient-based back-propagation training to provide symbolic solutions for differential equations that can be easily and directly used and interpreted. Building on the single-variable model shown in \cite{panju2020neuro}, we here focus on partial differential equations over multiple variables, as these problems are more common and more complex than differential equations over a single variable. Indeed, ordinary differential equations can be solved as a special case using our proposed method, along with other symbolic mathematical tasks such as integration, function inversion, and symbolic regression. As an added benefit, when our method fails to obtain an exact solution, including when no elementary solution exists, it will return a symbolic function that approximates the true solution, rather than leave you empty-handed.

In the sections that follow, we outline the core of our framework and the architecture of the multivariate symbolic function learner, or MSFL. We then show several experiments that demonstrate the power of our method on practical examples of PDEs. 
\section{Related work}
\label{rel_work}

There are several papers that apply neural networks for solving both ordinary and partial differential equations, including those by \cite{meade1994solution, lagaris1998artificial, berg2018unified, malik2020learning}. These approaches typically do not provide symbolic solutions to differential equations, but highly accurate approximate functions. There are also interesting works for discovering PDEs, such as by \cite{long2019pde}, which is a different problem from what we are addressing here.

One excellent work that directly links deep learning with symbolic mathematics is given by \cite{lample2019deep}. This paper attempts to train a transformer model to learn the language of symbolic mathematics and ``translate'' between inputs and outputs of mathematical tasks, such as integration and solving differential equations. Although the results are remarkably impressive, they depend upon an extremely costly training procedure that cannot scale well to new configurations, for example, involving a different set of operations than the ones used in the training set. On a deeper level, the model has faced criticism for issues ranging from the artificiality of its testing to the fact that it ``has no understanding of the significance of an integral or a derivative or even a function or a number'' \cite{davis2019use}.

The method and results presented in this work are an extension and development of the results in \cite{panju2020neuro}, which introduces a Symbolic Function Learner that can be used to solve or find approximate symbolic solutions to ordinary differential equations. In this work, we extend the method to apply to multivariate partial differential equations.

\section{Method}
\label{others}

There are two components to our symbolic PDE solution system. The multivariate symbolic function learning algorithm is a submodule that produces symbolic mathematical expressions over many variables in order to minimize a given cost function. The equation-solving wrapper is a system that uses the submodule to find solutions to partial differential equations. We will start by describing the wrapper system and how it relates to PDEs, and then outline a possible framework for the symbolic function learner that it uses.

\subsection{Equation Solver System}

A general form for representing any partial differential equation (PDE) is 
\begin{equation}
\label{g_defn}
    g\left(x_1, x_2, \ldots, x_d, y, \frac{\partial y}{\partial x_1}, \frac{\partial y}{\partial x_2}, \ldots, \frac{\partial y}{\partial x_d}, \frac{\partial^2 y}{\partial x_1^2}, \frac{\partial^2 y}{\partial x_1 \partial x_2}, \ldots\right) = 0.
\end{equation}

Here, $g$ is a real-valued mathematical expression in terms of the variables $x_1, \dots, x_d$, a function over these variables $y$, and any partial derivative of any order of $y$ over any set of the variables. 

It should be noted that this format encompasses far more than just PDEs; in fact,  integration, function inversion, and many functional equation problems can be expressed in the form of Equation \ref{g_defn}.

The solution to Equation \ref{g_defn} is the symbolic function $f(x_1, \ldots, x_d)$ that minimizes the expected error
\begin{equation}
\label{loss1}
    L_1(f) = \mathbb{E}\left[\left| g\left(x_1, \ldots, x_d, f(x_1, \ldots, x_d), \frac{\partial f}{\partial x_1}, \ldots, \frac{\partial f}{\partial x_d}, \frac{\partial^2 f}{\partial x_1^2}, \frac{\partial^2 f}{\partial x_1 \partial x_2}, \ldots \right) \right|^2\right]
\end{equation}
where $x_1, \ldots, x_d$ are distributed over the desired domain of $f$. Note that in most cases, it is sufficient to use discrete approximations for partial derivatives, such as
$$ \frac{\partial f}{\partial x_i} \approx \frac{f(\textbf{\textit{x}}+\textbf{\textit{e}}^{(i)}\varepsilon/2) - f(\textbf{\textit{x}}-\textbf{\textit{e}}^{(i)}\varepsilon/2)}{\varepsilon}$$
$$ \frac{\partial^2 f}{\partial x_i^2} \approx \frac{f(\textbf{\textit{x}}+\textbf{\textit{e}}^{(i)}\varepsilon) - 2f(\textbf{\textit{x}}) + f(\textbf{\textit{x}}-\textbf{\textit{e}}^{(i)}\varepsilon)}{\varepsilon^2}$$
% $$ \frac{\partial^2 f}{\partial x_ix_j} \approx \frac{f(x+\varepsilon(\textbf{\textit{e}}^{(i)} + \textbf{\textit{e}}^{(j)})) - f(x+\varepsilon(\textbf{\textit{e}}^{(i)} - \textbf{\textit{e}}^{(j)})) - f(x+\varepsilon(\textbf{\textit{e}}^{(j)} - \textbf{\textit{e}}^{(i)}))+ f(x-\varepsilon(\textbf{\textit{e}}^{(i)} + \textbf{\textit{e}}^{(j)}))}{4\varepsilon^2}$$
where $\textbf{\textit{e}}^{(i)}$ is the unit vector in the $i$th direction, and $\varepsilon$ is some small constant.

PDEs are frequently accompanied by boundary conditions or initial value constraints which solutions must satisfy. Often, these constraints come in the form of an interval, such as $f(x, 0) = c(x)$ for some specified $c$. We will approximate these constraints by taking $N$ uniformly spaced points $\left\{(\textbf{\textit{x}}_i, y_i)\right\}$ along the interval and inserting them into the secondary error function
\begin{equation}
    L_2(f) = \sum_i \left\| f(\textbf{\textit{x}}_i) - y_i\right\|^2.
\end{equation}

Combining these two error functions gives the total error 
\begin{equation}
\label{loss_eqn}
L_{total} = L_1(f) + \lambda L_2(f). 
\end{equation}
The best solution to the PDE is the function $f$ that minimizes $L_{total}.$ To find such an $f$, we must perform an optimization search using $L_{total}$ as an objective function. While neural networks do this naturally using back-propagation, it is not immediately clear how to perform this optimization over all symbolic functions. In the next section, we outline one way in which this can be done.

\subsection{Multivariate Symbolic Function Learner (MSFL)}
\label{explain_fl}

\begin{figure*}[t]
\centering
\includegraphics[width=0.95\textwidth]{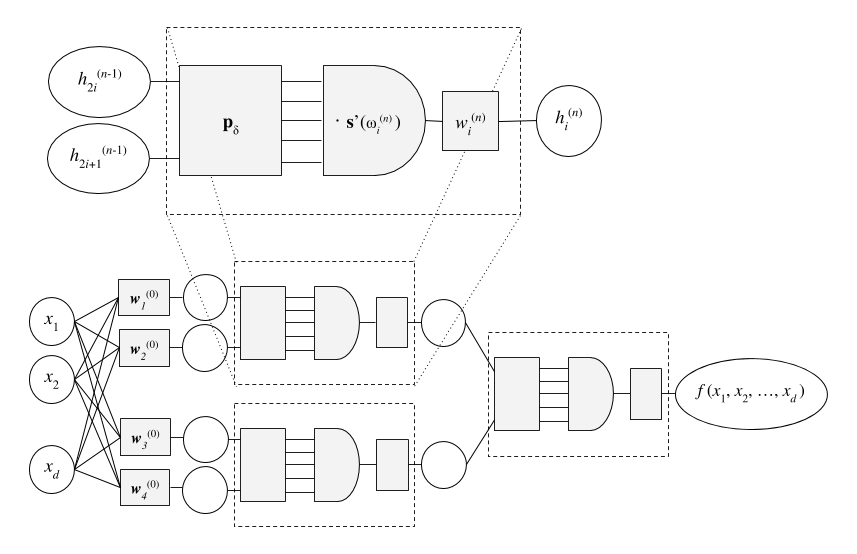} % Reduce the figure size so that it is slightly narrower than the column.
\caption{The architecture of the multivariate symbolic function learner (MSFL).  \textbf{Top:} The neural representation of a single operator (interior) node in the symbolic parse tree. The operator function $\textbf{p}$ takes in two child node as input and applies all available operators on their values. The discretized softmax function $\textbf{s}'$ is a gate that allows exactly one of these operators to pass through, determined by learnable weight $\omega$. This output is then scaled by learnable weight $w$. (Note: bias scalars $b$ are omitted from the diagram to save space.) \textbf{Bottom:} After an initial layer of leaf nodes which combine in a fully connected fashion, a series of operator nodes form the template of a balanced binary parse tree. The weight parameters determine how to interpret this tree as a single, well-formed symbolic mathematical expression $f$ over multiple variables.  }
\label{architecture}
\end{figure*}

The multivariate symbolic function learner (or MSFL for short) is an algorithm that generates symbolic mathematical expressions that minimize a given cost function. Any algorithm meeting this requirement will be a suitable fit for the proposed symbolic PDE solving system. In particular, many symbolic regression algorithms would fit the role with minor adjustments, including those by \cite{schmidt2009distilling},  \cite{sahoo2018learning}, and \cite{udrescu2020ai}. In this section, we describe one model designed for this task.

Recall that every mathematical expression can be represented as a syntactic parse tree, where nodes in the interior represent operators and leaf nodes represent variables or constants. To evaluate the represented expression, each operator takes in the values of its children nodes as input. By introducing an identity operator and defining unary operators to apply on a pre-specified combination of two child nodes, these parse trees can be standardized to be balanced and perfectly binary.

If the structure of a parse tree is taken as a template, then it is possible to produce a mathematical function by identifying what operations and input quantities to use in each node of the tree. This is the essence of the MSFL algorithm. The goal is to learn what each node represents in the expression that minimizes the given cost function. We will show how this can be done in a way that is fully differentiable, and hence caters to deep learning techniques, such as back-propagation.

% First, we consider the case where the functions are over a single real variable $x$.
Let $U$ be a list of allowable unary operators $[u_1, \ldots, u_r]$ that map $\mathbb{R}$ to $\mathbb{R}$, and let $V$ be a list of binary operators $[v_{r+1}, \ldots, v_k]$ that map $\mathbb{R}^2$ to $\mathbb{R}$, for a total of $k$ allowable operators. We define the ``operate'' function $\textbf{p}: \mathbb{R}^2 \rightarrow \mathbb{R}^k$ by 
\begin{eqnarray*}
\textbf{p}(x_1, x_2) &=& [ u_1(x_1 + x_2), \ldots, u_r(x_1 + x_2), v_{r+1}(x_1, x_2), \ldots, v_k(x_1, x_2) ].
\end{eqnarray*}

For example, if $U = [id, \sin, \exp{}]$ and $V =  [\times]$, then 
$$\textbf{p}(x_1, x_2) = [x_1+x_2, \sin(x_1+x_2), e^{x_1+x_2}, x_1x_2 ]. $$
Note that we use $id$ to refer to the unary identity function that returns its input value unchanged. In fact, it is equivalent to the addition operator because of how unary operators are defined to combine two child node inputs into one by using their sum.

We compute the $\textbf{p}$ function in each interior node of the parse tree. In this way, we are applying all possible operators to the node's two children as input. In order to interpret the parse tree as a mathematical expression, one of these operators should be selected to pass on its output value, to the exclusion of all others. This can be done using a trainable gating operation.

The gate can be set up as follows. Let $\omega$ be a learnable weight vector in $\mathbb{R}^k$. Denote by \textbf{s} the softmax function, i.e.
$$ \textbf{s}(\omega) = \left[\frac{e^{\omega_i}} {\sum_{j=1}^k e^{\omega_j}} \right]_{i=1}^k.$$

Now $ \textbf{s}(\omega)$ is a nonnegative vector in $\mathbb{R}^k$ with entries summing to 1. The dot product $\textbf{p}(h_1, h_2) \cdot \textbf{s}(\omega)$ is then a convex linear combination of the outputs over all operators allowed by $\textbf{p}$, skewed to give most weight to the operator at the index corresponding to the largest entry of $\omega$. The choice of operator that is represented by a given node can thus be learned by updating the learnable weight vector $\omega$ during training.

Although the softmax gate places most weight on a single entry of the vector of outputs passing through, it still retains nonzero contributions from all other operators in the output of $\textbf{p}$. This can be corrected by adjusting the output of $\textbf{s}(\omega)$ to become $\textbf{s}'(\omega)$, where $\textbf{s}'$ is a discretized form of softmax that returns a vector with 1 at the entry corresponding to the largest value of $\omega$  and 0 at  all other entries. One way to compute the discretized softmax is
$$\textbf{s}'(\omega) = \left[H_1\left(\ \frac{\textbf{s}(\omega)_i}{\max \textbf{s}(\omega) }\right)\right]_{i=1}^k $$
where $H_1$ is a narrow hump function centred at 1, such as $H_1(x) = e^{-1000(x-1)^2}$. Since the division operator, maximum function, and hump function are all differentiable almost everywhere, the discretized softmax preserves the differentiability of our model that allows deep learning using gradient-based training methods.

The above framework for a single operator node forms the foundational unit for the larger parse tree structure. Let $m$ represent the number of layers in the entire parse tree. The number of layers is a measure of how complex the represented mathematical expression is allowed to be. Trees with more layers form a richer space of mathematical functions, but provide a challenge by expanding the search space significantly.

For $i = 0, \ldots, 2^{m}-1$, define 
$$h_i^{(0)} = \textbf{w}_i^{(0)} \cdot \textbf{x} + b_i^{(0)}$$
where $\textbf{x}=[x_1, \ldots, x_d]$ is a vector of the variable in the mathematical expression being constructed, and $\textbf{w}_i^{(0)} \in \mathbb{R}^d$ and $b_i^{(0)} \in \mathbb{R}$ are learnable parameters.  This represents the lowest layer of the tree, consisting of leaf nodes that denote numerical quantities. Each of these quantities is in the form of a learnable linear combination of all possible variables.

Working up the layers of the tree, as $n = 1, \ldots, m$, the value of each node in recursively defined as
$$h_i^{(n)} = w_i^{(n)} \left(\textbf{s}'\left(\omega_i^{(n)}\right) \cdot \textbf{p}_\delta\left(h_{2i}^{(n-1)}, h_{2i+1}^{(n-1)})\right)  \right) + b_i^{(n)}$$
for each $i = 0, \ldots, 2^{m-n-1}$. Here, each $\omega_i^{(n)} \in \mathbb{R}^k$ and $w_i^{(n)}, b_i^{(n)} \in \mathbb{R}$ are learnable weights whose values will be learned during the training process.
 
The value at the root node of the tree, $h_0^{(m)}$, is the value of the mathematical expression represented by the tree when evaluated using the input $\textbf{x}$. It is by using this value for $\hat{f}(\textbf{x})$ that the MSFL can be trained using the cost function in Equation~\ref{loss_eqn}. Note that $\hat{f}(\textbf{x}) = h_0^{(m)}$ is obtained as a differentiable function over $\textbf{x}$ and all learnable weights in the MSFL.

At the end of the training procedure, the MSFL returns the symbolic expression represented by the final state of the parse tree. That is, it interprets each interior tree node as the operator determined by the weight in its $\omega$ vector, and applies an affine transformation to each node using the defined by the corresponding $w$ and $b$ values. If the resulting function is not the exact solution to the PDE, it will usually be a close approximation, as it is the result of a training procedure designed to optimize for a low fitting error.

\section{Experiments}

 We test our system on a number of PDE problems and demonstrate the results below. 

In each problem, we run our system over a function $g$ in the form of Equation~\ref{g_defn}. The MSFL algorithm is run using the unary operators $U = [id, \sin, \exp{}]$ and the binary operator $V =[\times]$. In order to avoid illegal argument errors, we automatically compute the absolute value before entering any value as input to operations defined only on the positive half-plane, such as the square root function. As mentioned in Section~\ref{explain_fl}, the nature of the $id$ operator removes the need for an explicit addition operator.

For each task, we run our method 20 times, training a model on 5000 randomly generated points within the domain of $f$ each time. Each run  returns the function represented by the parse tree after 6000 iterations. The standard softmax function $\textbf{s}$ is used for the first 1250 (= 25\%) of iterations of each run, and the discrete form $\textbf{s}'$ is used for the remainder of the training. This allows an introductory exploratory training phase before the model converges to a single expression structure, spending the remaining iterations fine-tuning the values of constants. After 20 runs are complete, we select from the 20 returned functions the one with the lowest validation error, as per Equation~\ref{loss_eqn}.

% \td{fix}
% We use a TensorFlow 2.2.0 implementation on Python 3.6, run on Microsoft Windows Server 2016 Standard OS with four Intel (R) Xeon (R) Gold 6126 CPU @ 2.60 GHz, 2594 Mhz, 12 core processors.

In all of the solution visualizations shown below, the left-most graphs are plots of the learned symbolic functions. The central graphs show the residual error from the learned functions, that is, the value of $g$ evaluated at $\textbf{\textit{x}}$ as in $L_1(f)$ used in Equation \ref{loss1}. The right graphs demonstrate how the learned functions fit the specified boundary conditions.
 
\subsection{Wave Equation}

The motion of wave travelling in one spatial dimension over time can be modelled by a function $u(x, t)$ that satisfies the PDE
\begin{equation}
\label{wave_eqn}
  \frac{\partial^2 u}{\partial t^2}(x, t) = c^2\frac{\partial^2 u}{\partial x^2 }(x, t)  
\end{equation}
for $x \in [0, \pi]$, $t > 0$.
This type of motion can be found in particles vibrating around a rest position along a single direction \cite{speiser2008discovering}.

Consider the case where $c=1$ and we are given the boundary conditions
\begin{eqnarray*}
u(0, t)=0 \\
u(\pi, t)= 0
\end{eqnarray*}
for $t > 0$, and initial conditions
\begin{eqnarray*}
u(x, 0) &=& 0 \\
\frac{\partial u}{\partial t}(x, 0) &=& \sin x
\end{eqnarray*}
for $x \in [0, \pi]$.

\begin{figure}[h]
\centering
\includegraphics[width=1.0\textwidth]{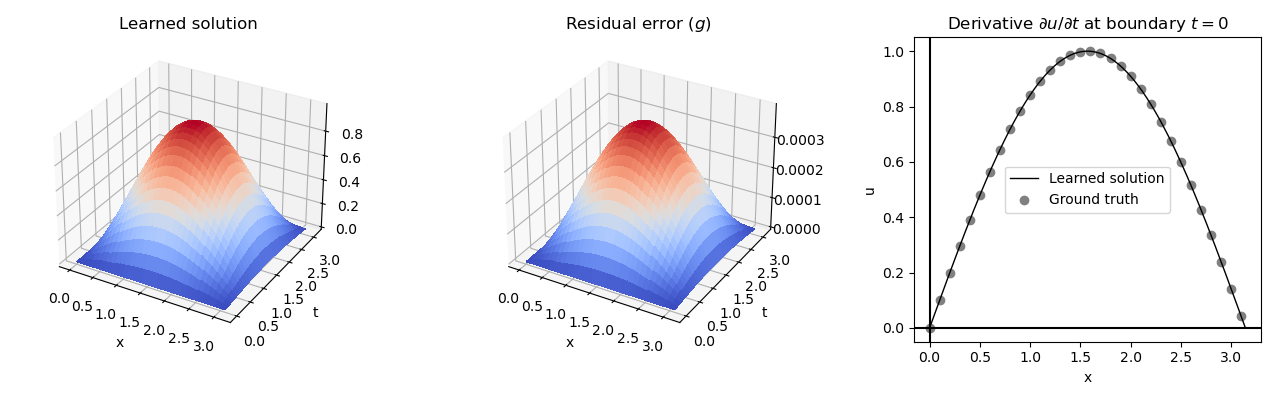} 
\caption{The learned solution to the wave equation (\ref{wave_eqn}), with residual error and boundary values.}
\label{wave_graph}
\end{figure}

This system has the exact solution $u(x, t) = \sin x \sin t.$
Our method produced the symbolic solution
$$\hat{u}(x, t) = 1.0002\sin(1.000x)\sin(0.9998t).$$
This result is represented visually in Figure~\ref{wave_graph}.

\subsection{Heat Equation}

The diffusion of heat through a medium along a single spatial dimension over time can be modelled by a function $u(x, t)$ that satisfies the PDE
\begin{equation}
\label{heat_eqn}
     \frac{\partial u}{\partial t} = \frac{\partial^2 u}{\partial x^2}
\end{equation} 
for $x \in [0, \pi]$, $t > 0$. For example, this equation might be used to model how an uneven initial distribution of heat disperses and levels out along a rod of uniform composition.

Consider the case given by the boundary conditions
\begin{eqnarray*}
u(0, t) = 0 \\
u(\pi, t) = 0
\end{eqnarray*}
for $t > 0$, and initial conditions
\begin{eqnarray*}
u(x, 0) = \sin x
\end{eqnarray*}
for $x \in [0, \pi]$.

\begin{figure*}[h]
\centering
\includegraphics[width=1.0\textwidth]{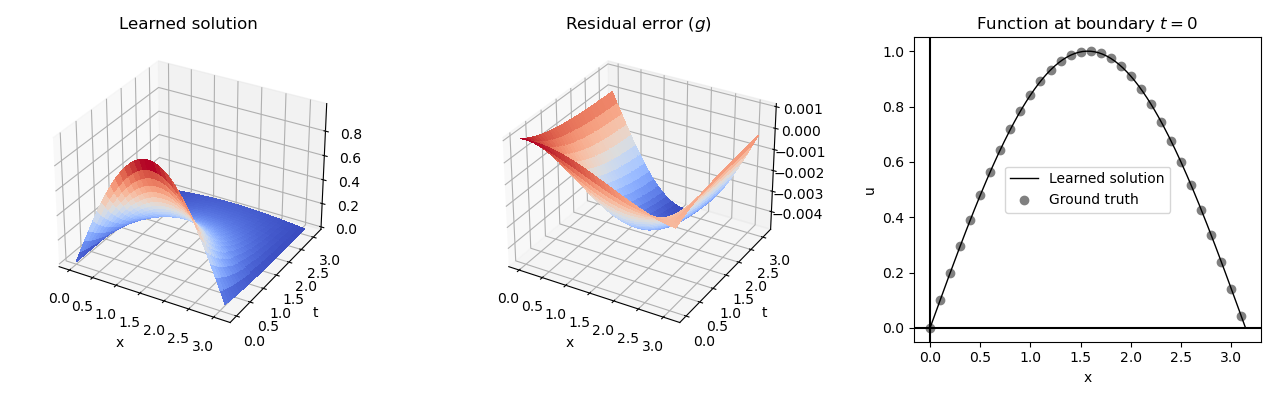} % Reduce the figure size so that it is slightly narrower than the column.
\caption{The symbolic solution to the heat equation (\ref{heat_eqn}), with residual error and boundary values.}
\label{heat_graph}
\end{figure*}

This system has the exact solution $u(x, t) = e^{-t} \sin{x}.$
Our method produced the symbolic solution
$$\hat{u}(x, t) = (1.005e^{-0.994t} - 0.005)\sin(0.9996x + 0.001t).$$
This result is represented visually in Figure~\ref{heat_graph}.

\subsection{Fokker-Planck Equations}

The Fokker-Planck equation is a general model describing the behaviour of particles undergoing Brownian motion, with applications ranging from astrophysics to economics \cite{risken1996fokker}. The general format of a linear Fokker-Planck (FP) equation, also called a forward Kolmogorov equation, is
\begin{equation}
\label{fpe_general}
    \frac{\partial}{\partial t} u(x, t) =  -\frac{\partial}{\partial x} A(x, t)u(x, t) 
+ \frac{\partial^2}{\partial x^2} B(x, t)u(x, t)
\end{equation} 
with initial condition $u(x, 0) = f(x)$, where $A(x,t)$ and  $B(x,t)$ (known as drift and diffusion coefficients, respectively), along with $f(x)$, are specified real-valued functions.

In order to fit the structure of Equation~\ref{g_defn}, we can reformulate Equation~\ref{fpe_general} as
\begin{equation}
    g\left(u, \frac{\partial u}{\partial t}, \frac{\partial x}{\partial t}, \frac{\partial^2 x}{\partial t^2}\right) = \frac{\partial u}{\partial t} - \left( 
    \left(\frac{\partial^2 B}{\partial x^2} -\frac{\partial A}{\partial x}\right)u
    +\left(2\frac{\partial B}{\partial x} - A\right)\frac{\partial u}{\partial x}
    +B\frac{\partial^2 u}{\partial x^2}
    \right) = 0.
\end{equation}

An alternative version for an FP equation, known as the backward Kolmogorov equation, is
\begin{equation}
\label{fpe_backwards}
    \frac{\partial}{\partial t} u(x, t) =  -A(x, t)\frac{\partial}{\partial x} u(x, t) 
+ B(x, t)\frac{\partial^2}{\partial x^2} u(x, t)
\end{equation} 
with initial condition $u(x, 0) = f(x)$, where $A(x,t),$ $B(x,t)$, and $f(x)$ are specified real-valued functions. Note that this formulation is already close to matching the format of Equation~\ref{g_defn}, and does not need to be significantly rearranged.

We will look at three common examples of FP equations with known analytic solutions. These examples have been investigated in several existing works on FP equations, including those by \cite{lakestani2009numerical}, \cite{dehghan2006use}, and \cite{kazem2012radial}.

\subsubsection{Example 1}

Consider the case of Equation~\ref{fpe_general} over $x, t \in [0, 1]$ where 
$ A(x, t) = -1$ and $ B(x, t) = 1$,
subject to the initial condition $f(x) = x$.

\begin{figure}[h]
\centering
\includegraphics[width=1.0\textwidth]{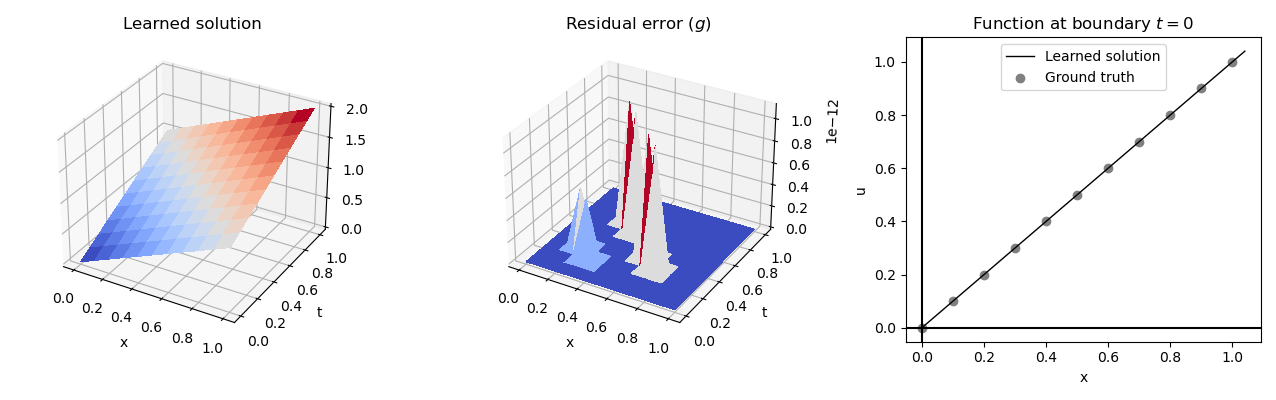} 
\caption{The symbolic solution to FP Example 1, with residual error and boundary values.}
\label{fpr1_graph}
\end{figure}

This system has the exact solution $u(x, t) = x + t.$
Our method produced the symbolic solution
$$\hat{u}(x, t) = 1.0001x + 1.0001t.$$
This result is represented visually in Figure~\ref{fpr1_graph}.

\subsubsection{Example 2}

Consider the case of Equation~\ref{fpe_general} over $x, t \in [0, 1]$ where $ A(x, t) = x$ and
$B(x, t) = x^2/2$,
subject to the initial condition $f(x) = x$.

\begin{figure}[h]
\centering
\includegraphics[width=1.0\textwidth]{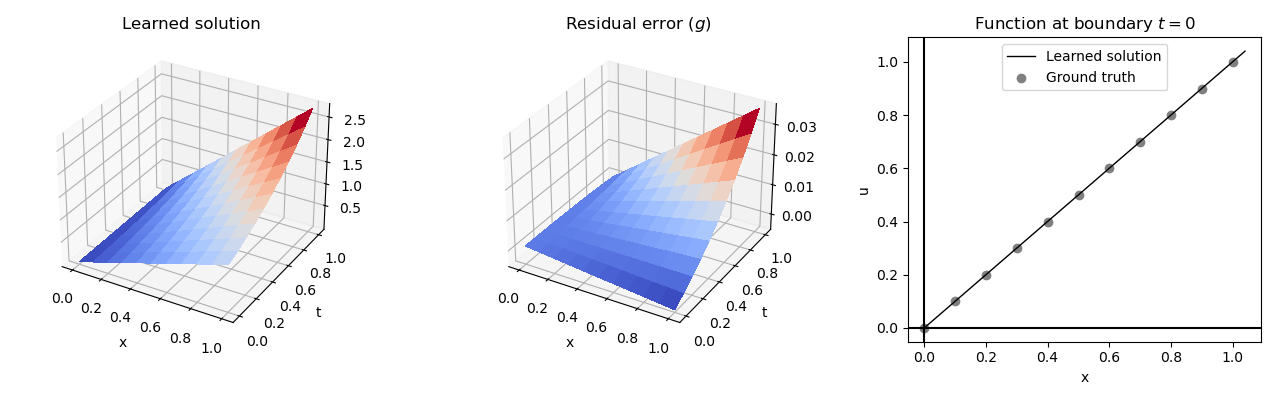} 
\caption{The symbolic solution to FP Example 2, with residual error and boundary values.}
\label{fpr2_graph}
\end{figure}

This system has the exact solution $u(x, t) = xe^t.$
Our method produced the symbolic solution
$$\hat{u}(x, t) = (0.972x - 0.001t + 0.002)e^{1.024t} + 0.029x - 0.002.$$
This result is represented visually in Figure~\ref{fpr2_graph}.

\subsubsection{Example 3}

Consider the case of Equation~\ref{fpe_backwards} over $x, t \in [0, 1]$ where 
$ A(x, t) = -(x+1)$ and 
$ B(x, t) = x^2 e^t$,
subject to the initial condition $f(x) = x+1$.

\begin{figure}[h]
\centering
\includegraphics[width=1.0\textwidth]{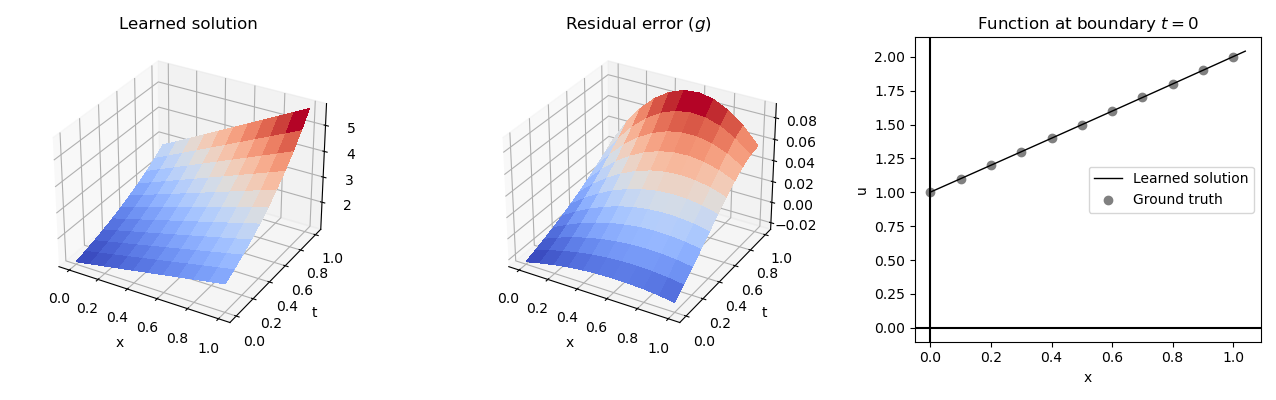} 
\caption{The symbolic solution to FP Example 3, with residual error and boundary values.}
\label{fpr3_graph}
\end{figure}

This system has the exact solution $u(x, t) = (x + 1)e^t.$
Our method produced the symbolic solution
$$\hat{u}(x, t) = (0.923x - 0.006t + 0.866)e^{0.011x + 1.121t} + 0.057x + 0.136.$$
This result is represented visually in Figure~\ref{fpr3_graph}.

\section{Conclusions and future work}
In this work, we have shown a framework for producing symbolic solutions to partial differential equations over many variables using deep learning techniques. We have illustrated the utility of our system on a number of examples of PDEs taken from classical physics. Our method has shown its ability to generate solutions that are either exactly or approximately correct in many cases. In particular, the linear Fokker-Planck equations have provided good testing grounds for the strengths of our method.

Although the multivariate symbolic function learner has shown its capability, there is still room for improvement. As with all algorithms that seek to optimze over symbolic functions, the problem of an immensely vast search space poses a great challenge. It would be rewarding to see an MSFL sucessfully scale to spaces of increasingly complex functions over large sets of operators and variables. The modularity of our system allows the MSFL to be easily swapped with any other function learning algorithm, offering a good opportunity for future experiments.

In the bigger picture, we look forward to seeing more applications of deep learning in the realm of symbolic mathematics, and hope that this contribution will be a step towards that direction.

\bibliography{iclr2021_conference}
\bibliographystyle{plain}

% \appendix
% \section{Appendix}
% You may include other additional sections here.

\end{document}